\documentclass{article}
\usepackage{spconf,amsmath,graphicx,hyperref,booktabs,amssymb,multirow}

\title{Explainable Deepfake Detection with RL Enhanced Self-Blended Images}
\name{\begin{tabular}{@{}c@{}}
		Ning Jiang\textsuperscript{1},
		Dingheng Zeng\textsuperscript{2}, 
		Yanhong Liu\textsuperscript{2},
		Haiyang Yi\textsuperscript{2}, 
		Shijie Yu\textsuperscript{2}, \\
		Minghe Weng\textsuperscript{2},  
		Haifeng Shen\textsuperscript{2}, 
		and Ying Li\textsuperscript{1*}\thanks{\small{© 2026 IEEE. Personal use of this material is permitted. Permission from IEEE must be obtained for all other uses, in any current or future media, including reprinting/republishing this material for advertising or promotional purposes, creating new collective works, for resale or redistribution to servers or lists, or reuse of any copyrighted component of this work in other works.}}
		\end{tabular}
		\vspace{-0.15in}}
\address{\textsuperscript{1}School of Software \& Microelectronics, Peking University, Beijing, China\\
\textsuperscript{2}Mashang Consumer Finance Co., Ltd., Chongqing, China\vspace{-0.15in}}

\begin{document}
%

\maketitle
\begin{abstract}
Most prior deepfake detection methods lack explainable outputs. With the growing interest in multimodal large language models (MLLMs), researchers have started exploring their use in interpretable deepfake detection. However, a major obstacle in applying MLLMs to this task is the scarcity of high-quality datasets with detailed forgery attribution annotations, as textual annotation is both costly and challenging—particularly for high-fidelity forged images or videos. Moreover, multiple studies have shown that reinforcement learning (RL) can substantially enhance performance in visual tasks, especially in improving cross-domain generalization. To facilitate the adoption of mainstream MLLM frameworks in deepfake detection with reduced annotation cost, and to investigate the potential of RL in this context, we propose an automated Chain-of-Thought (CoT) data generation framework based on Self-Blended Images, along with an RL-enhanced deepfake detection framework. Extensive experiments validate the effectiveness of our CoT data construction pipeline, tailored reward mechanism, and feedback-driven synthetic data generation approach. Our method achieves performance competitive with state-of-the-art (SOTA) approaches across multiple cross-dataset benchmarks. Implementation details are available at \href{}{https://github.com/deon1219/rlsbi.}
\end{abstract}
\vspace{-0.05in}
\section{Introduction}
\vspace{-0.05in}
\label{sec:intro}
The rapid advancement of deepfake techniques presents significant challenges to existing face forgery detection systems, which have primarily focused on improving generalization to unknown forgery types \cite{li2020face,cao2022end,yan2023ucf,yan2024transcending}. Recent developments employ MLLMs for both detection and explainability \cite{guo2025rethinkingvisionlanguagemodelface,yu2025unlockingcapabilitieslargevisionlanguage}, while annotated datasets with authenticity labels and CoT annotations \cite{zhang2024commonsensereasoningdeepfake,qin2025interactivedeepfakeanalysis} enhance model interpretability and generalization.

However, their detection performance still lags behind specialized models. This gap primarily stems from the inherent difficulty in accurately describing forgery clues via CoT annotations, particularly for high-quality fabricated media. Current methodologies often rely on labor-intensive processes involving human annotators and GPT-4o-assisted comparisons between real and fake images to localize forgery regions and generate CoT annotations. However, these approaches remain constrained by high data requirements and potential inaccuracies in forgery localization \cite{chen2025x2dfdframeworkexplainableextendable}.

Building on the demonstrated potential of Group Relative Policy Optimization (GRPO) in Deepseek-R1 \cite{deepseekai2025deepseekr1incentivizingreasoningcapability}, R1-inspired reinforcement learning has shown promise in multimodal vision tasks \cite{huang2025visionr1incentivizingreasoningcapability,shen2025vlm}. However, applying R1 to face forgery detection - fundamentally a binary classification task - remains challenging due to reward signal sparsity.

To address this, we propose a novel framework integrating blending techniques \cite{shiohara2022sbi} with R1, using precise forgery-conditioned keywords as reward signals. This represents the first successful application of R1 in face forgery detection, effectively mitigating reward sparsity while improving both accuracy and generalization to novel forgery methods. Our key contributions are:

\noindent 1) We propose an automatic framework for precise forgery description generation. Our method first extracts generation conditions from forged images, then maps them to forgery regions/clues, which are expanded into CoT data with MLLMs. This approach significantly mitigates hallucination effects while eliminating the need for extensive manual verification required by prior methods.

\noindent 2) We introduce a keyword-driven reward mechanism specifically tailored for forgery detection. By embedding high-level semantic keywords of forgery clues into CoT data and implementing reward functions to constrain model predictions, our method effectively addresses the reward sparsity challenge in GRPO training for binary classification tasks.

\noindent 3) We develop an adaptive feedback mechanism integrated with GRPO training. This system dynamically adjusts generation strategies based on historical reward values, enabling progressive learning through iterative optimization. 
\begin{figure*}[!t]
	\centering
	\includegraphics[width=.95\textwidth]{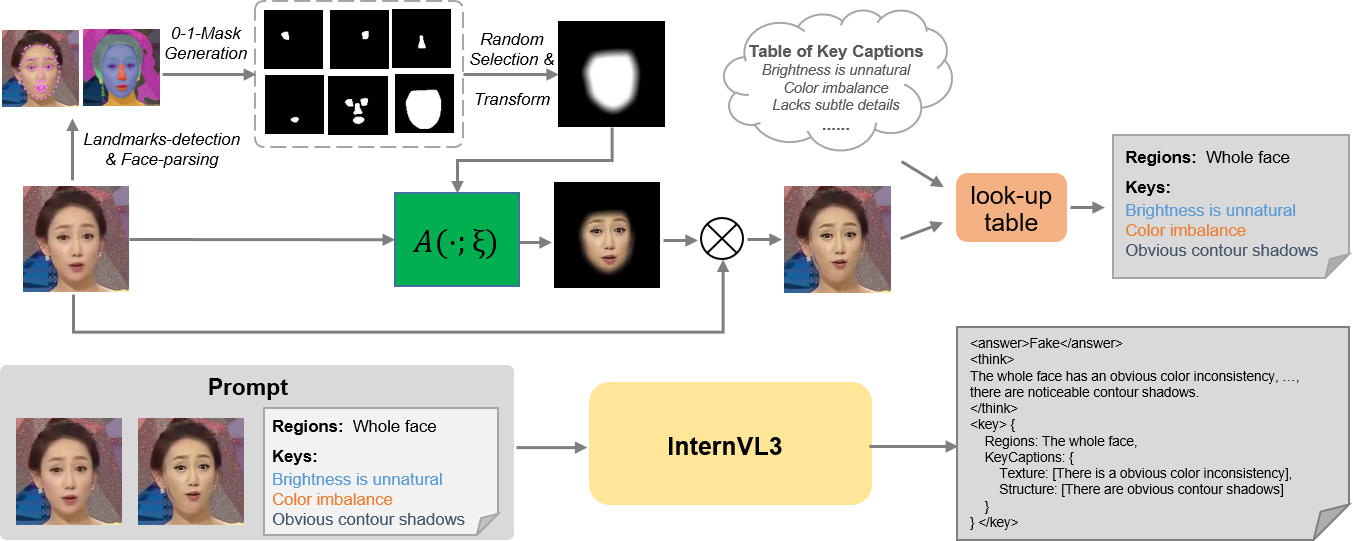} 
	\caption{The pipeline of generating CoT data. First, using a randomly selected mask and perturbation parameters to generate a fake image. Then retrieve key captions from the lookup table based on the mask and parameters. Finally, feed the image pair and key captions into an MLLM to produce the corresponding CoT annotation.}
	\vspace*{-.2in}
	\label{fig:sbi_data}
\end{figure*}
\vspace{-0.05in}
\section{Related Work}
\label{sec:related}
\vspace{-.05in}
{\noindent\bf Deepfake detection.} Recent deepfake detection work using MLLMs incorporate specialized detector structures for improving performance \cite{peng2025mllmenhancedfaceforgerydetection,chen2025x2dfdframeworkexplainableextendable} or leverage external knowledge from the forgery image generation process to generate CoT data \cite{he2025vlforgeryfacetriaddetection}.  FFTG~\cite{vlffd2025} locates forged regions and generates simple textual descriptions by performing pixel-level comparisons between real images and their corresponding forged counterparts. However, FFTG requires applying deepfake techniques to generate forged counterparts for real images, which is time-consuming and labor-intensive. The quality of the generated descriptions heavily depends on the diversity of the constructed forgery methods and images.
Our approach builds upon SBI~\cite{shiohara2022sbi} and operates solely on real samples, dynamically generating forged data online. This allows us to precisely identify forged regions, obtain segmentation masks, and simultaneously generate both textual descriptions and fully accurate forensic clues about the forgery.

{\noindent\bf Reinforcement learning.} The R1-like reinforcement learning methodology has been applied to MLLM research~\cite{feng2025videor1reinforcingvideoreasoning,peng2025lmmr1empowering3blmms}. Curr-ReFT~\cite{deng2025boostinggeneralizationreasoningvision} introduces curriculum learning approaches that progress from simple to complex tasks. Vision-R1~\cite{huang2025visionr1incentivizingreasoningcapability} constructs high-quality mathematical reasoning datasets through human and GPT4o annotations to guide reinforcement training. VLM-R1~\cite{shen2025vlm} further analyzed the critical role of reward function design in enhancing MLLM capabilities. 
\vspace{-0.2in}
\section{Method}
\vspace{-0.05in}
\label{sec:method}
\subsection{Construction of Basic Forged Data}
\vspace{-.05in}
Inspired by SBI \cite{shiohara2022sbi}, we propose synthesizing fake images using the real data from FaceForensics++'s training set \cite{rossler2019faceforensics}. As shown in Figure~\ref{fig:sbi_data}, the process begins by detecting 81-point landmarks on a real face, followed by face parsing to segment the facial regions (full face, left eye, right eye, nose, and mouth). These segmented parts are then randomly combined to generate forgery region masks, which include 15 single-organ combinations and 4 predefined combinations (full face, eyebrows, and forehead).

A series of augmentation operations are applied to the input real image, with randomly eroded/dilated mask regions. The augmentation considers multiple factors: hue, lighting, clarity, contrast, scaling, and translation, each applied at three intensity levels (mild, moderate, and severe). Finally, the augmented image is blended with the original face to produce the fabricated data. Mathematically, the forged image $\mathcal{I}_f$ is generated as:\vspace*{-.08in}
$$\mathcal{I}_f = \alpha \cdot \mathcal{T}(\mathcal{I}_m) \odot \mathcal{A}(\mathcal{I}_r;\xi)  + (1-\alpha \cdot \mathcal{T}(\mathcal{I}_m)) \odot \mathcal{I}_r\vspace*{-.08in}$$
where $\alpha$ is the blending weight, $I_r$ is the real image, $\mathcal{T}$ denotes the mask transformation, $\mathcal{A}(\cdot;\xi)$ represents the augmentation operation where $\xi$ is the randomly selected parameters, and $\mathcal{I}_m$ denotes the forged mask region.
\vspace{-.1in}
\subsection{Construction of CoT Data}
\vspace{-.05in}
We define a ‌key caption table‌ that indicates editing regions and discrepancy descriptions caused by modifications. Anomaly keywords for the forged image are randomly selected from this table, based on the measured difference thresholds between the forged and original images.

Using the generated forged images and their synthesis conditions, we employ the open-source multimodal model ‌InternVL3-38B\cite{zhu2025internvl3exploringadvancedtraining}‌ to produce reasoning descriptions in CoT format‌. Given the actual tampered regions and associated anomaly descriptions, we input these into the model. Through ‌prompt engineering‌, we guide the multimodal model to autonomously select the most contextually appropriate description while ensuring semantic distinctiveness.
\vspace{-.1in}
\subsection{Text-Based Forgery Localization Reward}
\vspace{-.05in}
In addition to the accuracy and format rewards in GRPO's framework, we introduce a task-specific reward signal for deepfake detection: text-based forgery localization reward. By adopting SBI-like methods with controllable forgery generation, we obtain precise forgery region information that can be infinitely generated online. This deterministic positional description serves as one of the model's reward signals.

During training, we randomly generate position masks and convert them into textual descriptions. From the model's text output, we extract the content corresponding to the {\em Regions} field in the key tags. Let ${REG}_{pred}$ and ${REG}_{gt}$ denote the sets of regional fields for the model response and ground truth, respectively. The forgery localization reward is then calculated using the Jaccard similarity between these two sets. 
\vspace{-.15in}
\subsection{Model Tuning}
\vspace{-.05in}
To validate the effectiveness of our proposed CoT data construction method and the enhancement of reinforcement learning in deepfake detection tasks, we follow a traditional two-stage training approach.
\vspace{-.1in}
\subsubsection{Supervised Fine-Tuning}\vspace{-.05in}
Using offline-generated CoT data, we conduct Supervised Fine-Tuning (SFT) on the MLLM in a Visual Question Answering (VQA) format. The objective is to enable the model to follow the required format, extract key anomalies from images, describe forgery clues, and provide final conclusions.
In this phase, we adopt a hybrid training strategy combining Faceforensics++ (FF++) data  without CoT descriptions and SBI data with CoT descriptions. FF++ data encompasses four common deepfake manipulation types. By converting binary labels into ``Real/Fake'' textual annotations, the model can learn effective features for distinguishing genuine from forged images.
Furthermore, by generating forged data using SBI methods along with corresponding forgery clue descriptions, the model establishes associations between forged features and textual descriptions, ensuring outputs adhered to the specified format.
Through this training strategy, we eliminated the need for a dedicated forgery detector. The model achieved robust binary classification capabilities and interpretable textual outputs in a single training step.
\vspace{-.1in}
\subsubsection{Reinforcement Training}\vspace{-.05in}
We perform GRPO on the SFT model. Similar to the SFT phase, we use a mix of FF++ data without CoT descriptions, and the SBI data generated online. 
\begin{figure}[!t]
	\includegraphics[width=.485\textwidth]{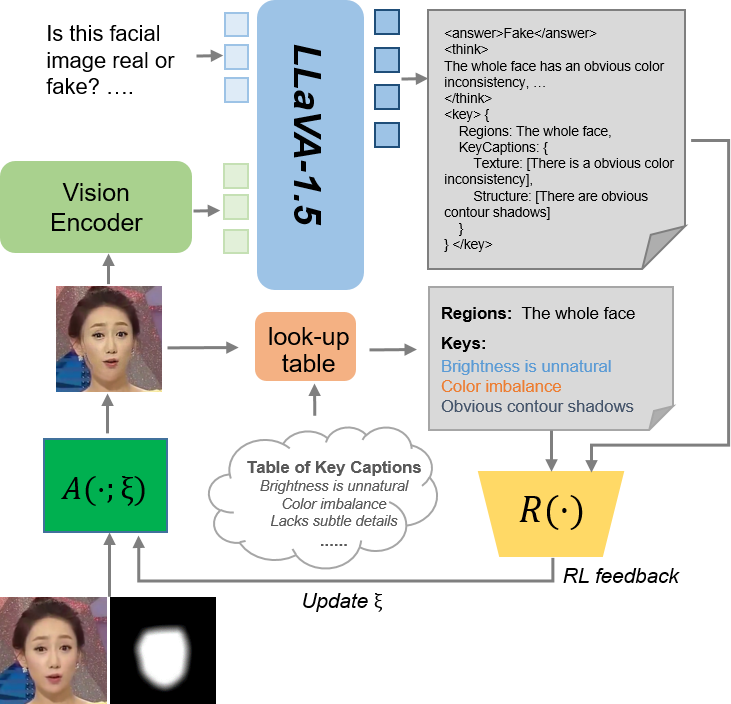} 
	\caption{The feedback-based GRPO training. Dynamically adjust the data generation parameters based on computed historical reward values.}
	\vspace*{-.2in}
	\label{fig:rl}
\end{figure}

During GRPO training, we calculate the following rewards: classification reward $R_{acc}$ (extracting real/fake labels and comparing with ground truth), format reward $R_{format}$ (verifying adherence to the format structure as specified in the prompt), keyword reward $R_{key}$ (measuring the ROUGE score between extracted forgery clues in \textless key\textgreater...\textless /key\textgreater and ground truth keywords, which includes the forgery localization reward calculating overlap between forged region descriptions and ground truth annotations),length reward $R_{len}$ (constraining response length within predefined bounds). The total reward score is obtained by the sum of each item:
$$\vspace{-0.1in}R_{total}= R_{acc}+R_{format}+R_{key}+R_{len}\vspace{-0.01in}$$

Additionally, we adjust the generation parameters based on the mean/stability of historical reward values, to generate harder fake data with lower rewards and easier fake samples vice versa. Through GRPO reinforcement training, the model learns from the reward function's feedback, demonstrating improved generalization to unseen forged data compared to the SFT baseline.
\vspace{-0.05in}
\section{Experiment}
\vspace{-.05in}	
\label{sec:exp}

{\noindent\bf Dataset}. Following works like \cite{yan2025orthogonalsubspacedecompositiongeneralizable}, we used preprocessed data from DeepfakeBench \cite{DeepfakeBench_YAN_NEURIPS2023} for training and testing. The training set includes the FF++ and fake data (with corresponding text annotationss) generated via SBI from FF++'s real samples. The test set comprised of CDF2 \cite{li2020celebdf}, DFD \cite{google2019deepfake}, DFDC \cite{dolhansky2020dfdc}, and DFDCP \cite{dolhansky2019preview} datasets.

{\noindent\bf Evaluation Metrics}. To align with prior works, we report AUC as the primary metric. Since the model outputs text (unable to compute AUC directly from Answer's Real/Fake labels), we adopted the method from $X^2$-DFD to derive AUC and EER by calculating answer probabilities. 

{\noindent\bf Implementation Details}. Consistent with $X^2$-DFD and M2F2-Det \cite{guo2025rethinking}, we employed LLaVA 1.5-7b as the base model and leveraged Video-R1's framework \cite{feng2025videor1reinforcingvideoreasoning} for SFT and RL training. 
During ‌SFT training, LoRA was applied to ViT and LLM (Rank=8, Alpha=16, dropout=0.1), with the projector trainable.
We set the learning rate, batch size, gradient accumulation steps to $2e-4$, 4 and 2 respectively. The model was trained for 10 epochs (last epoch used for evaluation and RL baseline).

We used full fine-tuning for RL training. The learning rate and batch size is set to $1e-6$ and 3 respectively. Following Video-R1, GRPO was implemented via vLLM (8 generations, beta=0.04, 1 iteration). We used 4× NVIDIA A800 GPUs and trained for 1000 steps. Further training incurred instability due to sharpened action distributions and tightened KL-penalties amplifing gradient noise from dynamic feedback.

{\noindent\bf Main Results}. We compared SOTA methods in deepfake detection and reported frame-level and video-level results in Table \ref{tab:res-frame-level} and Table \ref{tab:res-video-level}, respectively. In the table headers, ``DFD'' refers to dedicated deepfake detectors, while ``MLLM'' indicates that the method employs a multimodal large language model framework. If both DFD and MLLM are checked, it means the method incorporates a dedicated deepfake detector as an auxiliary component alongside the base MLLM framework. Our method uses only the base MLLM framework without introducing any additional detector.
\begin{table}[!htbp]
	\centering
	\caption{Frame-level cross-dataset evaluation using AUC metric. For each target dataset, the highest performance is shown in bold, the second-best result is underlined.}
	\label{tab:res-frame-level}
	\setlength\tabcolsep{2.5pt}
	\begin{tabular}{lcccccc}
		\toprule
		Method & DFD & MLLM & CDF2 & DFDC & DFDCP & DFD \\
		\midrule
		X-Ray~\cite{li2020face} & \checkmark & & 0.679 & 0.633 & 0.694 & 0.766 \\
		RECCE~\cite{cao2022end} & \checkmark & & 0.732 & 0.713 & 0.734 & 0.812 \\
		SBI~\cite{shiohara2022sbi} & \checkmark & & 0.813 & - & 0.799 & 0.774 \\
		UCF~\cite{yan2023ucf} & \checkmark & & 0.753 & 0.719 & 0.759 & 0.807 \\
		ED~\cite{ba2024exposing} & \checkmark & & 0.864 & 0.721 & 0.851 & - \\
		LSDA~\cite{yan2024transcending} & \checkmark & & 0.830 & 0.736 & 0.815 & 0.880 \\
		ProDet \cite{cheng2024can} & \checkmark & & 0.842 & - & 0.774 & 0.848 \\
		Trident \cite{kara2025tridentdetectingfaceforgeries} & \checkmark & & 0.861 & 0.826 & 0.845 & 0.920 \\
		F. Adap. \cite{cui2025forensicsadapterunleashingclip} & \checkmark & & 0.900 & \textbf{0.843} & \underline{0.890} & \underline{0.933} \\
		$X^2$-DFD & \checkmark & \checkmark & \underline{0.903} & \underline{0.835} &\textbf{0.897} & 0.925 \\
		FFTG \cite{vlffd2025}& & \checkmark & 0.832 & - & 0.832 & \textbf{0.948} \\
		RLSBI (Ours) & & \checkmark & \textbf{0.905} & 0.818 & 0.817 & 0.926 \\
		\bottomrule
	\end{tabular}
\end{table}

Table \ref{tab:res-frame-level} presents cross-dataset evaluation results. Data for other methods are sourced from the Forensics Adapter, $X^2$-DFD, and FFTG papers. Our method outperforms previous approaches on CDF2, significantly surpassing FFTG, which shares a similar framework. On the large-scale DFD dataset, our method also exceeds most DFD and MLLM-based methods.

Table \ref{tab:res-video-level} displays video-level AUC results. On CDF2, our method ranks second only to VLF-FFD, which incorporates a dedicated deepfake detector, and performs comparably to SOTA methods on DFD. However, on the DFDC and DFDCP datasets, our method trails behind SOTA approaches. We attribute this performance gap to the intrinsic mismatch between SBI’s focus on blending artifacts and DFDC’s diverse, non-blending forgery types coupled with severe environmental degradations.

\begin{table}[!t]
\vspace{-.2in}
\centering
\caption{Video-level cross-dataset evaluation using AUC metric, trained with FF++ c23 dataset.}
\label{tab:res-video-level}
\setlength\tabcolsep{2.pt}
\begin{tabular}{lcccccc}
	\toprule
	Method & DFD & MLLM & CDF2 & DFDC & DFDCP & DFD \\
	\midrule
	RECCE~\cite{cao2022end} & \checkmark & & 0.823 & 0.696 & 0.734 & 0.891 \\
	SBI~\cite{shiohara2022sbi} & \checkmark & & 0.886 & 0.717 & 0.848 & 0.827 \\
	UCF~\cite{yan2023ucf} & \checkmark & & 0.837 & 0.742 & 0.770 & 0.867 \\
	LSDA~\cite{yan2024transcending} & \checkmark & & 0.875 & 0.701 & 0.812 & 0.881 \\
	ProDet \cite{cheng2024can} & \checkmark & & 0.926 & 0.707 & 0.828 & 0.901 \\
	CDFA \cite{lin2024fake} & \checkmark & & 0.938 & 0.830 & 0.881 & 0.954 \\
	F. Adap. \cite{cui2025forensicsadapterunleashingclip} & \checkmark & & 0.957 & \underline{0.872} & \textbf{0.929} & -- \\
	Effort \cite{yan2025orthogonalsubspacedecompositiongeneralizable} & \checkmark & & 0.956 & 0.843 & 0.909 & 0.965 \\
	$X^2$-DFD~\cite{chen2025x2dfdframeworkexplainableextendable} & \checkmark & \checkmark & 0.955 & 0.853 & 0.912 & 0.957 \\
	KFD \cite{yu2025unlockingcapabilitieslargevisionlanguage}& \checkmark & \checkmark & 0.947 & 0.791 & 0.918 & \textbf{0.996} \\
	M2F2-det \cite{guo2025rethinking}& \checkmark & \checkmark & 0.951 & \textbf{0.878} & -- & \underline{0.977} \\
	VLF-FFD \cite{peng2025mllmenhancedfaceforgerydetection} & \checkmark & \checkmark & \textbf{0.973} & 0.854 & \underline{0.920} & -- \\
	RLSBI (Ours) & & \checkmark & \underline{0.963} & 0.839 & 0.849 & 0.965 \\
	\bottomrule
\end{tabular}
\vspace{-.2in}
\end{table}

{\noindent\bf Ablation Studies}. Table \ref{tab:ablation} presents the ablation studies conducted at the frame level, validating the effectiveness of each component in our RL training strategy. The baseline model (SFT), after undergoing simple instruction tuning, already demonstrates certain detection capabilities. The introduction of a reward mechanism based on key evidence verification (+key\_verify) leads to a modest performance improvement, confirming that our designed reward function effectively guides the model to focus on critical semantic features in forged regions. With the incorporation of the feedback-guided (\text{+feedback}) online data synthesis strategy, the model exhibits a significant performance gain on the CDF2 dataset, with the AUC increasing from 0.882 to 0.905 and the EER decreasing from 0.204 to 0.176.

\begin{table}[htbp]
	\vspace{-.2in}
	\centering
	\caption{Ablation study on the RL strategy. Performance is evaluated using frame-level AUC and EER metrics.}
	\label{tab:ablation}
	\setlength\tabcolsep{5.2pt}
	\begin{tabular}{llccccc}
		\toprule
		Phase & Method & \multicolumn{2}{c}{CDF2} & \multicolumn{2}{c}{DFDCP} \\
		\cmidrule(lr){3-4} \cmidrule(lr){5-6}
		& & AUC & EER & AUC & EER \\
		\midrule
		SFT & FF++/SBI\_CoT & 0.881 & 0.206 & 0.804 & 0.275 \\
		\cline{1-6}  
		\multirow{2}{*}{RL} & +key\_verify & 0.882 & 0.204 & 0.815 & \textbf{0.264} \\
		& +feedback & \textbf{0.905} & \textbf{0.176} & \textbf{0.817} & 0.265 \\
		\bottomrule
	\end{tabular}
	\vspace{-.2in}
\end{table}
\vspace{-0.05in}
\section{Conclusion}
\vspace{-.1in}
In this paper, we propose a method that leverages self-blending image techniques to generate forged images along with accurate textual descriptions of the forgeries, aiming to address the scarcity of high-quality text-annotated data required by current MLLM-based deepfake detection systems. Furthermore, we introduce a curriculum learning training strategy guided by reinforcement feedback to synthesize data, which enhances the model's cross-domain detection performance to SOTA levels without incorporating additional dedicated deepfake detectors, while also enabling the model to produce precise descriptions of deepfake clues. We also believe that the use of reinforcement feedback to expand data distributions offers promising avenues for further research and exploration.

\renewcommand{\baselinestretch}{0.}\small\normalsize
\bibliographystyle{IEEEbib}
\bibliography{refs-abr}

\end{document}